\documentclass[lettersize,journal]{IEEEtran}
\usepackage{amsmath,amsfonts}
\usepackage{algorithmic}
\usepackage{algorithm}
\usepackage{array}
\usepackage[caption=false,font=normalsize,labelfont=sf,textfont=sf]{subfig}
\usepackage{textcomp}
\usepackage{stfloats}
\usepackage{url}
\usepackage{verbatim}
\usepackage{graphicx}
\usepackage{cite}
\usepackage{booktabs}
\usepackage{multirow}
\usepackage{algorithm}
\usepackage{algorithmic}
\usepackage{amsmath}
\usepackage{mathtools}
\usepackage{amsfonts}
\usepackage{booktabs}
\usepackage{arydshln}

\begin{document}

\title{Joint Combinatorial Node Selection
and Resource Allocations in the Lightning Network using Attention-based Reinforcement Learning}

\author{Mahdi Salahshour\textsuperscript{*}, Amirahmad Shafiee\textsuperscript{*}, Mojtaba Tefagh\\
\small{\textsuperscript{*}These authors contributed equally.}
\thanks{Mahdi Salahshour, Amirahmad Shafiee, and Mojtaba Tefagh are with the Department of Mathematical Sciences, Sharif University of Technology, Iran (e-mail: mahdi.salah@sharif.edu; amirahmad.shafiee@sharif.edu; mtefagh@sharif.edu)}

}



\maketitle

\begin{abstract}
The Lightning Network (LN) has emerged as a second-layer solution to Bitcoin's scalability challenges. The rise of Payment Channel Networks (PCNs) and their specific mechanisms incentivize individuals to join the network for profit-making opportunities. According to the latest statistics, the total value locked within the Lightning Network is approximately \$500 million\footnote{https://defillama.com/protocol/lightning-network}. Meanwhile, Joining the LN with the profit-making incentives presents several obstacles, as it involves solving a complex combinatorial problem that encompasses both discrete and continuous control variables related to node selection and resource allocation, respectively. Current research inadequately captures the critical role of resource allocation and lacks realistic simulations of the LN routing mechanism. In this paper, we propose a Deep Reinforcement Learning (DRL) framework, enhanced by the power of transformers, to address the Joint Combinatorial Node Selection and Resource Allocation (JCNSRA) problem. We have improved upon an existing environment by introducing modules that enhance its routing mechanism, thereby narrowing the gap with the actual LN routing system and ensuring compatibility with the JCNSRA problem. We compare our model against several baselines and heuristics, demonstrating its superior performance across various settings. Additionally, we address concerns regarding centralization in the LN by deploying our agent within the network and monitoring the centrality measures of the evolved graph. Our findings suggest not only an absence of conflict between LN’s decentralization goals and individuals' revenue-maximization incentives but also a positive association between the two. 
\end{abstract}


\section{Introduction}
The Lightning Network (LN) \cite{lightning} is a Payment Channel Network (PCN) layered on top of the Bitcoin network \cite{bitcoin}, serving as a decentralized second-layer solution to enhance scalability, performance, and privacy by offloading the majority of transactions from the blockchain \cite{tikhomirov2020quantitative}. In LN, payments are source-routed through the least costly paths, motivating routing nodes to strategically position themselves on as many cost-effective routes as possible \cite{how-to-profit}. Consequently, an essential task for a node operator is to optimize channel opening and capacity allocation strategies to maximize expected routing opportunities and, in turn, increase revenue.

LN's structure and routing incentives are influenced by strategic behavior, where nodes aim to maximize routing opportunities and minimize costs by selecting optimal connections and fee policies. This strategic positioning reflects rational economic behavior, as nodes seek to maximize revenue \cite{carotti2024rational, bertucci2020incentives}. 
Current efforts to maximize revenue for a single node in the LN primarily focus on optimizing fee policies \cite{Aida-Kiana, how-to-profit, zhang2023payment} or selecting optimal nodes for channel openings \cite{lightning-gym}. However, these approaches often neglect the crucial role of channel capacity allocation, a key component in LN routing mechanisms \cite{hubs}, which is essential for maximizing revenue.

The problem of Joint Combinatorial Node Selection and Resource Allocation (JCNSRA) involves a series of intertwined continuous-discrete subproblems. The complexity arises from the discrete combinatorial aspects, akin to those in well-known problems like combinatorial multi-armed bandits \cite{chen2014combinatorial} and combinatorial feature selection \cite{charikar2000combinatorial}, as well as the continuous challenges found in distinguished problems like resource allocation \cite{hurwicz1973design}, compounded by the additional difficulty of integrating these elements. Moreover, we examine the JCNSRA problem on a large-scale, real-world dynamic graph with active flow, further compounding the complexity of understanding the underlying nature of the problem.

In this paper, we formulate the primary problem as a Markov Decision Process (MDP), enabling the agent to arrive at the optimal solution in a sequential manner. We leverage the capabilities of Reinforcement Learning (RL) and transformers \cite{attention} to address this complex and challenging problem.
To effectively solve the JCNSRA problem in LN, it is crucial to have an environment with a high-quality routing mechanism. We introduce our simulator, a thorough and multi-configurable JCNSRA environment featuring a realistic routing mechanism by dynamically adjusting node channels' liquidity through transaction flow streams. 

We further examine one of the concerns within the LN, decentralization. The primary goal of the LN is to create a decentralized platform that
scales the Bitcoin network and supports high-throughput micropayments \cite{lightning}. However, the network’s evolution and changes in its topology have faced scrutiny and criticism for its growing centralization \cite{pili2021topological}. We deploy our developed agents into the LN snapshot and monitor its impact on the centralization of the evolved network.

The primary contributions of this paper are as follows:
\begin{itemize}

    \item We developed our new simulator, a comprehensive gym environment with a routing mechanism improved by online liquidity updates, tailored to the JCNSRA problem. Built on a preexisting environment, our simulator introduces a localization method, providing an extensive empirical platform for model evaluation across various scales and configurations, which we leverage to assess our proposed method.
    
    \item We propose a novel node selection and capacity allocation module for the LN that leverages the capabilities of transformers and the Proximal Policy Optimization (PPO) \cite{ppo} algorithm. This approach significantly outperforms common heuristics and baseline methods in solving this complex combinatorial task.
    
    \item We examine the LN centralization concern by deploying our agent to analyze the topological evolution of the LN and measure different centralities. We observe a positive influence of our economically rational agents on the network's decentralization.
    
\end{itemize}

The structure of this paper is as follows: we begin with a discussion of essential backgrounds and the related work in the literature in section \ref{sec:background}, followed by a formulation of the JCNSRA problem in section \ref{sec:formulation}. Next, in sections \ref{sec:approach} \& \ref{sec:architecture}, we describe our problem within an RL framework and introduce the details of our environment, along with our approach, which leverages DRL and transformers. We then deploy our proposed agent on our LN snapshot to analyze its evolution and share the details in section \ref{sec:analysis}. Subsequently, we discuss the results, demonstrating how our model outperforms existing baselines and heuristics in section \ref{sec:results}. Finally, we present a brief discussion and summarize our findings in the conclusion.

\section{Background \& Related Works}\label{sec:background}
\subsection{Background}

\subsubsection*{Lightning Network} \label{LN-appendix}
Bitcoin is a decentralized digital payment system operating on a peer-to-peer network without a central authority \cite{bitcoin}. A Bitcoin transaction, known as an on-chain transaction, allows parties to exchange Bitcoin, which must be recorded on the blockchain. As previously mentioned, scalability is a significant challenge for blockchain systems, leading to the development of PCNs. LN, a prominent PCN, can be represented as a graph where nodes represent participants connected by channels that maintain liquidity balances. These channels are the fundamental mechanism for value exchange within the LN, enabling off-chain payments that do not require blockchain registration.


In LN, channels between participants are established through an on-chain transaction known as a funding transaction. Once this transaction is confirmed on the blockchain, the channel is opened, and its capacity is set. This capacity defines the upper limit of transaction amounts that can be sent through the channel and is initiated with the sum of provided funding by both parties involved. Participants also generate a commitment transaction after each routing, which is not broadcast to the blockchain but reflects the current liquidity balance between them. This liquidity is divided into two components for each party: inbound liquidity refers to the capacity a participant has to receive funds through a payment channel, while outbound liquidity indicates the capacity to send funds. In a channel between two participants, the inbound liquidity of one participant is equal to the outbound liquidity of the other, as the total capacity of the channel is shared between both parties. As transactions occur over the LN, the sender’s outbound liquidity decreases, while the recipient’s outbound liquidity increases correspondingly. This design enables an arbitrary number of off-chain, feeless transactions between two parties to be effectively consolidated into just two on-chain transactions: one for opening the channel and one for closing it. When the parties decide to close the channel, they submit the last updated commitment transaction to the blockchain
\cite{lightning, hubs}.

To minimize the number of channels required for transactions, LN incorporates multi-hop routing, which facilitates payments between non-adjacent nodes. In this process, payments traverse intermediate nodes along a path preselected by the sender. Although LN utilizes multiple routing strategies \cite{emergent, hoenisch2018aodv, prihodko2016flare}, the sender is typically assumed to determine the optimal route using a modified version of Dijkstra’s shortest path algorithm \cite{dijkstra2022note, vsatcs2020understanding}. This algorithm is tailored to account for channel capacities, fees, and lock durations when calculating edge weights. Specifically, the algorithm first eliminates candidate edges that lack sufficient capacity and then selects the path with the lowest cumulative edge weights, factoring in the fee structures and maximum lock times set by the intermediate nodes \cite{mccorry2016towards}.

For calculating the associated transaction fees, each edge in the public network graph stores the routing fee policies, which consist of a base fee and a proportional fee. The base fee is a fixed charge that must be paid to the routing node for every payment forwarded. The proportional fee, on the other hand, is a variable charge, calculated by multiplying the transaction amount by a predefined rate. As a result, higher-value payments bring forth higher fees for the routing nodes \cite{emergent}.\\

\subsubsection*{Reinforcement Learning}\label{RL-appendix} RL is a subfield of machine learning where an agent learns to interact with an environment to maximize cumulative rewards over time. This problem is commonly modeled as an MDP, represented by the tuple $(\mathcal{S}, \mathcal{A}, P, R, \gamma)$. Here, $\mathcal{S}$ denotes the set of possible states, $\mathcal{A}$ represents the set of actions available to the agent, $P(s' | s, a)$ is the state transition probability function that defines the likelihood of moving to state $s'$ after taking action $a$ in state $s$, $R(s, a)$ is the reward function that provides the scalar reward for taking action $a$ in state $s$, and $\gamma \in [0,1)$ is the discount factor that determines the present value of future rewards. \cite{rl}

The agent's behavior in the RL framework is governed by a policy $\pi$, which maps states to a probability distribution over actions, i.e. $\pi: \mathcal{S} \rightarrow \mathcal{P}(\mathcal{A})$. The objective of the agent is to learn an optimal policy that maximizes the expected cumulative reward, given by:
\begin{equation} G_t = \sum_{k=0}^{\infty} \gamma^k R(s_{t+k}, a_{t+k}) \end{equation}
 DRL enhances conventional RL by utilizing deep neural networks to approximate elements such as the policy. This approach allows agents to handle high-dimensional and complex state spaces more efficiently.

\subsubsection*{Proximal Policy Optimization}\label{PPO-appendix}PPO \cite{ppo} is a state-of-the-art policy gradient method that tries to find the optimal policy and addresses the challenges of instability and inefficient exploration in RL. 
PPO improves upon traditional policy gradient methods by introducing a clipped surrogate objective. The objective function that PPO maximizes is defined as:

\small
\begin{equation}
L(\theta) = \mathbb{E}_t\left[\min\left(r_t(\theta) \hat{A}_t, \text{clip}(r_t(\theta), 1 - \epsilon, 1 + \epsilon)\hat{A}_t\right)\right]
\end{equation}
\normalsize
where $r_t(\theta) = \frac{\pi_{\theta}(a_t | s_t)}{\pi_{\theta_{\text{old}}}(a_t | s_t)}$ is the probability ratio between the new policy $\pi_{\theta}$ and the old policy $\pi_{\theta_{\text{old}}}$, $\hat{A}_t$ is an estimator of the advantage function \cite{schulman2015high}, which indicates how much better action is compared to the average action at that state, and $\epsilon$ is a hyperparameter that controls the extent of clipping. The $\text{clip}(\cdot)$ function ensures that the probability ratio $r_t(\theta)$ stays within a specified range $[1 - \epsilon, 1 + \epsilon]$, preventing large policy updates that could destabilize training.\\

\subsubsection*{Transformer}\label{Transformer-appendix}The transformer \cite{attention} is an encoder-decoder, architecture built with the attention mechanism as its main core. The encoder module of the Transformer consists of $N$ identical transformer blocks. Each block consists of two sub-layers.


The first sub-layer is a multi-head attention mechanism. Each attention head in this mechanism performs scaled-dot-product attention. Given queries $\mathbf{Q}$, keys $\mathbf{K}$, and values $\mathbf{V}$, scaled-dot-product attention is computed as:
\begin{equation}
    \text{Attention}(\mathbf{Q}, \mathbf{K}, \mathbf{V}) = \text{softmax}\left(\frac{\mathbf{Q}\mathbf{K}^\top}{\sqrt{D_k}}\right)\mathbf{V} = \mathbf{A}\mathbf{V}
\end{equation}
where $\mathbf{A} = \text{softmax}\left(\frac{\mathbf{Q}\mathbf{K}^\top}{\sqrt{D_k}}\right)$ is the attention matrix. To enhance representational power, the Transformer employs multi-head attention by projecting $\mathbf{Q}$, $\mathbf{K}$, and $\mathbf{V}$ into $H$ different subspaces, applying scaled-dot-product attention in each subspace, and concatenating the results:

\begin{equation} 
    \text{MHAttn}(\mathbf{Q}, \mathbf{K}, \mathbf{V}) = \text{Concat}(\text{head}_1, \dots, \text{head}_H)\mathbf{W}^O
\end{equation}

where each head is computed as:
\begin{equation}      
    \text{head}_i = \text{Attention}(\mathbf{Q}\mathbf{W}_i^{Q}, \mathbf{K}\mathbf{W}_i^{K}, \mathbf{V}\mathbf{W}_i^{V})
\end{equation}
The second sub-layer is a position-wise fully connected feed-forward network. This network applies the same transformation independently to each position:
\begin{equation}
    \text{FFN}(\mathbf{H}') = \text{ReLU}(\mathbf{H}'\mathbf{W}^1 + \mathbf{b}^1)\mathbf{W}^2 + \mathbf{b}^2
\end{equation}
where $\mathbf{H}'$ is the output of the previous layer, $\mathbf{W}^1$ and $\mathbf{W}^2$ are weight matrices, and $\mathbf{b}^1$ and $\mathbf{b}^2$ are bias vectors.\\

Each sub-layer output is finalized by employing a residual connection followed by layer normalization. Specifically:
\begin{equation}
    \mathbf{H}' = \text{LayerNorm}(\text{SelfAttention}(\mathbf{X}) + \mathbf{X})
\end{equation}

\begin{equation}
    \mathbf{H} = \text{LayerNorm}(\text{FFN}(\mathbf{H}') + \mathbf{H}')
\end{equation}
where $\text{SelfAttention}(\cdot)$ refers to the multi-head self-attention mechanism and $\text{LayerNorm}(\cdot)$ denotes layer normalization.
We adopt the previous notations from this work \cite{lin2022survey}.\\

\subsubsection*{Graph Neural Networks}\label{GNN-appendix}
Graph Neural Networks (GNNs) \cite{kipf2016semi,velivckovic2017graph,hamilton2017inductive} are a class of neural network models specifically designed to learn representations of various components of a graph, such as its nodes, edges, and the graph as a whole. These models are capable of capturing the complex relationships and structural information embedded in graph data. At the core of GNNs is the message-passing scheme \cite{gilmer2017neural}, which iteratively updates the embedding of a node $v$ by incorporating information from its neighboring nodes. This process can be described as:
\begin{equation}
\begin{aligned}
\mathbf{h}_v^{(l)} &= \text{GNN}^{(l)} \left( \mathbf{h}_v^{(l-1)}, \left\{ \mathbf{h}_u^{(l-1)} \mid u \in \mathcal{N}(v) \right\}, \mathbf{A} \right) \\
&= \mathbf{C}^{(l)} \left( \mathbf{h}_v^{(l-1)}, \mathbf{A}^{(l)} \left( \left\{ \mathbf{h}_u^{(l-1)} \mid u \in \mathcal{N}(v) \right\} \right) \right)
\end{aligned}
\end{equation}

In this formulation, $\mathbf{h}_v^{(l)}$ denotes the embedding of node $v$ at layer $l$, where $l \in \{1, \dots, L\}$, and $\mathcal{N}(v)$ represents the set of $v$'s neighbors. The functions $\mathbf{A}^{(l)}$ and $\mathbf{C}^{(l)}$ correspond to the message aggregation and embedding update processes at the $l$-th layer, respectively. To obtain a representation of the entire graph, the embeddings of all nodes at the final layer $L$ are aggregated using a readout function, formally expressed as:
\begin{equation}
\mathbf{h}_G = \text{READOUT}\left(\left\{ \mathbf{h}_v^{(L)} \mid v \in V \right\}\right)
\end{equation}
where $\mathbf{h}_G$ is the resulting graph-level representation. The READOUT function can vary, often involving averaging or other pooling techniques appropriate to the graph structure \cite{lee2019self,ying2018hierarchical,zhang2018end}. We adopt the previous notations from this work \cite{ju2024survey}.

\subsection{Related Works}
Since the advent of the LN, the maximization of personal revenue has gained decent attention from researchers. Recent studies have primarily focused on channel-opening strategies and node-selection methodologies. For instance, \cite{lightning-gym} employs RL integrated with GNNs \cite{gnn}, while \cite{how-to-profit} proposes a greedy yet exhaustive algorithm for sequentially opening optimal channels. Both approaches, however, depend heavily on centrality measures, which may not be the most effective metrics given the intricate mechanics of LN's routing algorithms. Additionally, in \cite{Aida-Kiana} the authors explore revenue maximization through fee-setting policies, leveraging the PPO algorithm and utilizing reward signals derived from LN flow simulations. Nevertheless, these existing works exhibit significant limitations: they either lack realistic rewarding mechanisms or metrics \cite{lightning-gym, how-to-profit}, fail to generalize across different network topologies \cite{Aida-Kiana}, or rely on simplistic greedy algorithms \cite{how-to-profit}. Moreover, none of these studies adequately address the critical issue of capacity allocation, which plays a pivotal role in transaction throughput. In this paper, we aim to address all of these limitations comprehensively.

\section{Problem Formulation}\label{sec:formulation}
Consider a directed graph $G = (V, E)$, where each node $v \in V$ represents a participant in the LN and is associated with specific attributes. Each directed edge $e \in E$ corresponds to the channels between parties in the LN and is characterized by three attributes: ``base-fee'', ``fee-rate'' (proportional fee), and ``balance'' (outbound liquidity). Notably, an established channel between two parties is represented by a bidirectional edge between the two corresponding nodes in the network. Both participants can send payments through this channel corresponding to the balance attributes of their directed edge.We define a simulator function \( S(\cdot) \) such that \( S(G) = (V, E') \). Each edge \( e \in E' \) is augmented with a list of simulated transactions, which we denote as \( e_F \). For the sake of simplification, the flow is generated in three discrete amounts, based on observations on the LN transaction amount densities \cite{hubs}. Also, each transaction is routed based on a realistic balance updating mechanism adapted from the LN in our simulator \cite{lightning}.

\begin{figure}[tbp!]
     \centering
        {\includegraphics[width = 0.45\textwidth, keepaspectratio,  trim= 0cm 0cm 0.4cm 0cm, clip]{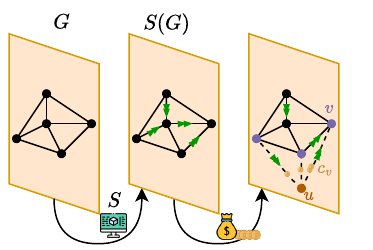}}

        \caption{Overview of flow simulation and the solution to the JCNSRA problem.}

        \label{fig:Problem}
\end{figure}

In the MDP formulation of the JCNSRA problem, as shown in Figure \ref{fig:Problem}, our objective is to select a list of tuples like $[(v_1, \tilde{c}_{v_1}), \dots, (v_T, \tilde{c}_{v_T})]$, respectively. Here,  $v_t \in V$ represents the selected node at timestep $t$ and each corresponding $\tilde{c}_{v_t}$ denotes an allocation of resources to the node $v_t$, where  $T$ represents the episode length. Note that each node $v \in V$ is initially allocated with $c_v = 0$. There are also no strict requirements for the distinctiveness of $v_t$; if multiple selected $v_t$ are similar, their allocated resources will naturally be aggregated. Based on all $v \in V$ with non-zero allocated resources through the episode and their corresponding aggregated resources $c_v = \sum_{v_t = v}{\tilde{c}_{v_t}}$ we establish new channels between our new attached node $u$ and the existing graph $S(G)$ and $E_u$ represents the set of edges between $u$ and the LN graph. Finally, our objective is to maximize the revenue gained from the imposed fees on transaction flows associated with the opened channels. Noteworthy, while a transaction passes through a node via both an inward and an outward channel, fees are charged only when receiving a transaction through an inward channel.

\section{Approach}\label{sec:approach}
The JCNSRA problem, focused on revenue maximization, is not only dependent on the network topology of the graph but also critically influenced by the flow streams and dynamics within the network. To effectively understand these graph flow dynamics, it is essential to grasp the underlying routing mechanisms. In the case of the LN, common transaction routing strategies follow a cheapest-path approach based on Dijkstra's algorithm \cite{traffic, cloth}. Comprehending the behavior of such a system is complex due to its reliance on unknown specific channel features and ambiguous transaction end-points. Consequently, it is imperative for the model to capture these network dynamics accurately \cite{meirom2021controlling}.

To effectively address these issues utilizing DRL algorithms, we have formulated the JCNSRA problem in the context of LN as an MDP, decomposing it into a multi-step problem of node selection and resource allocation. Due to the alignment of MDP with the formulation of dynamic problems, we expect that the model will be able to learn the hidden graph flow dynamics.

\subsection{State Space}
To effectively capture both the static and dynamic characteristics of the graph, we utilize the following features:
\begin{itemize}
    \item \emph{Degree Centrality:} This metric represents the degree of a node, which, in our context, is normalized.

    \item \emph{Provider Status:} In the LN, certain nodes serve as service providers, meaning they primarily function as payees, receiving payments for the services they offer \cite{lightning}. We hypothesize that being in close proximity to these provider nodes positively influences the volume of transaction flows through a node. Provider status is thus represented as a binary feature, where a value of 1 indicates a service provider node and a value of 0 indicates all other nodes.

    \item \emph{Transaction Flow:} At each timestep, a node passes a specific volume of transactions. These transaction volumes are normalized across all nodes to ensure comparability.

    \item \emph{Allocated Share:} At each timestep, nodes that have been previously selected by the agent hold a specific amount of resources. These resource amounts are normalized across all nodes for uniformity.
\end{itemize}

The state at each timestep $t$, denoted as $s_t$, is defined as
$s^t = [g^t_1, g^t_2, \dots, g^t_N]$ in which $g^t_i = [d_i, p_i, f^t_i, a^t_i]$ where $N = |V|$ represents the number of nodes in the graph, $g^t_i$ indicates the feature vector of the $i$th node at timestep $t$. Here, $d_i$ and $p_i$ correspond to the degree centrality and provider status of the $i$th node, respectively. Additionally, $f^t_i$ and $a^t_i$ represent the transaction flow and allocated share for the $i$th node at timestep $t$.

\subsection{Action Space}
As previously discussed, in the defined MDP framework, at each timestep $t$, the agent is required to select a node $v_t$ and allocate a resource share $\tilde{c}_{v_t}$. For further simplification, we require the agent to allocate a discrete share from the set $[1,\dots, K]$. The size of the resulting action space is $N \times K$, where $K$ is the size of the discrete allocation space. The action $a_t$ at timestep $t$ is thus defined as $a_t = (v_t, c'_{v_t})$, where $v_t \in V$ is the selected node, and $ c'_{v_t} \in [1,\dots, K]$ is the discrete resource allocated to that node at timestep $t$.

At each timestep $t$, for the already selected set of actions $[(v_1, c'_{v_1}),\dots,(v_t, c'_{v_t})]$, we first calculate the intended allocation values by normalizing $c'_{v_t}$s by their and then scale these allocation shares to the total allocation budget i.e $\tilde{c}_{v_t} = (\frac{c'_{v_t}}{\sum^t_{j=1} c'_{v_j}})C$ where $C$ is the total allocation budget. We will then calculate the allocation values $c_v$ for all $v \in V$ as described in section \ref{sec:formulation}.


Consider a scenario at timestep 3, with a total allocation budget of 100, $K = 10 $, and selected actions \([(a, 3), (b, 2), (a, 5)]\). In this case, \( c'_{v_1} = 3 \), \( c'_{v_2} = 2 \), and \( c'_{v_3} = 5 \), where \( v_1 = a \), \( v_2 = b \), and \( v_3 = a \). The allocation values are calculated as $\tilde{c}_{v_1} = \frac{3}{10} \cdot 100 = 30$, $\tilde{c}_{v_2} = \frac{2}{10} \cdot 100 = 20$, and $\tilde{c}_{v_3} = \frac{5}{10} \cdot 100 = 50$. Finally, the allocated resources for nodes \( a \) and \( b \) are determined by summing their respective values
$c_a = \sum_{v_t = a}{\tilde{c}_{v_t}} = 30 + 50 = 80$ and $c_b = \sum_{v_t = b}{\tilde{c}_{v_t}} = 20$.


In the proposed approach, the final solution to the JCNSRA problem hinges on both the selection of specific nodes and the efficient allocation of their resources. Solving the JCNSRA problem directly would necessitate exploring an action space of size $(N \times K)^T$, leading to exponential growth in potential actions as the number of channels increases. Instead, by framing the problem as an MDP, we can limit the action space to $N\times K$. This reduction effectively removes the exponential dependency on $T$, resulting in a more manageable action space that scales linearly with $N$ and $K$. Consequently, this reformulation makes the RL process more computationally feasible, supporting faster exploration and improved policy optimization, especially as the number of channel establishments grows.

\subsection{Reward Function}

As discussed earlier, our objective is to maximize our revenue from charged fees over transactions flowed by our inward-connected channels. Assuming the simulator function closely approximates the transaction flows that occur in LN, we hence maximize the reward function $R$, defined as the following:

\begin{equation}
R(G', u) = \sum_{e \in  E^{\text{in}}_u} \sum_{\emph{f} \in e_F} \emph{f} \cdot r(e) + b(e) 
\end{equation}

Here, \( G' = (V', E') \), where \( V' = V \cup \{u\} \) and \( E' = E \cup E_u \). Additionally, \( E^{\text{in}}_u \) denotes the incoming directed edges from other nodes to our node \( u \). The fee-rate for edge \( e \) is represented by \( r(e) \), and the base-fee is represented by \( b(e) \). The challenge is to determine the optimal configuration for establishing \( E_u \) in order to maximize the total reward \( R \), as defined above.

\section{Model Architecture}\label{sec:architecture}
\begin{figure*}[htbp!]
     \centering
        {\includegraphics[width=1\textwidth, keepaspectratio]{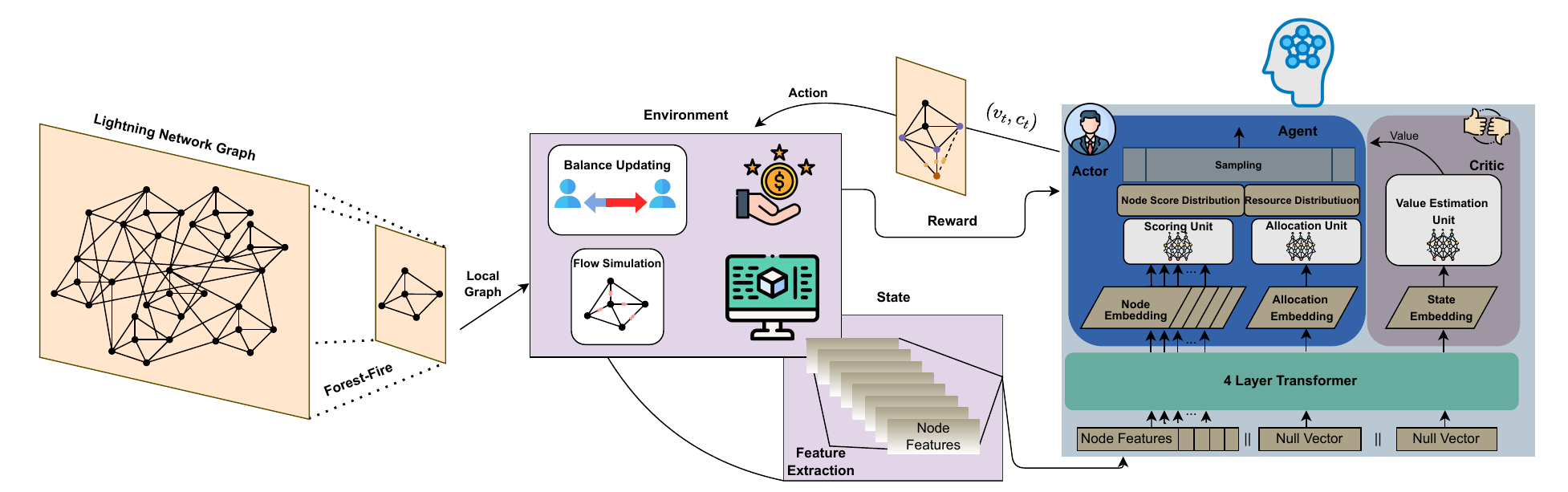}}
        \caption{Overview of the solution pipeline and model architecture.}

        \label{fig:Pipeline}
\end{figure*}

In this section, we will provide insights into our solution pipeline and model architecture. As illustrated in Figure \ref{fig:Pipeline}, we will start by describing our environment and its specific characteristics. Following up, we delve into our agent and its main decision-making unit's architecture.

\subsection{Environment}
Our experimental environment builds upon the foundational characteristics of a prior work \cite{Aida-Kiana}. We utilize a snapshot of the LN from 2021 \cite{snapshot}. These snapshots are created by executing a specific command on an LN node instance allowing any typical LN participant to generate their own snapshot. Each raw snapshot includes the details of public payment channels, but we focus on extracting only the key topological properties of the LN, including nodes, payment channels, channel fee policies, and channel capacities, along with supplementary data on service provider nodes derived from 1ML data\footnote{https://1ml.com/}. It is important to emphasize that any snapshot of the LN can be utilized, provided it retains these essential characteristics and adheres to a similar structural format.

Due to LN's stringent privacy policies, critical transaction details—such as sender, receiver, and transaction amounts—are not disclosed \cite{lightning}. Consistent with the methodology of the referenced environment, we simulate transactions randomly, with particular attention to the distribution of receivers, as service providers are statistically more likely to be recipients \cite{martinazzi2020evolving}.

We implemented several modifications to the environment, aimed at improving certain mechanisms and ensuring compatibility with the JCNSRA problem. These enhancements and adaptations are discussed in detail in the subsequent subsections.\\

\subsubsection*{Localization} Simulating the original LN graph presents notable challenges due to the vast number of nodes (roughly 16000 nodes) and the constraints of computational resources. Hence, the adoption of an effective localization strategy is crucial. Previous work employs a basic localization approach by deterministically aggregating nodes in proximity to the source node, starting with its immediate neighbors.

We enhanced this localization approach by incorporating the forest-fire sampling method \cite{fireforest}. The forest-fire method is among the most effective techniques for localizing large graphs, as it preserves both static and dynamic graph patterns inherent to the original network \cite{sampling}. By utilizing this method, the localized connected subgraph, despite variations in size, retains the structural properties of the overall network. Additionally, in each training episode, we generate a new localized graph randomly, ensuring that the agent can generalize across different graphs of similar size and develop robustness to local structural variations.\\

\subsubsection*{Flow dynamics} Liquidity dynamics, driven by transaction flows within the LN, play a crucial role in shaping the network's dynamic characteristics \cite{hubs}. This underscores the need for real-time updates to channel balances during the simulation process. To address this, we implemented per-transaction balance updates for all inbound and outbound channels along the sender-receiver path, thereby enhancing the fidelity of the simulation to more closely reflect the actual operational mechanisms of the LN.\\

\subsubsection*{Configuration Diversity}Our environment is designed to define the problem across a range of complexities and configurations. The proposed JCNSRA problem is constrained by several factors, including sample size, the number of channels, discrete allocation size, and budgetary limitations. Its compatibility with various configurations allows researchers to tailor their environment to match specific problem complexities and research objectives.

Notably, the integration of the Gym library \cite{gym}, along with its compatibility with the Stable-Baselines3 library \cite{sb3}, facilitates the straightforward application of diverse model architectures across different configurations of this problem.

\subsection{Agent} 

\begin{figure}
    \centering
    \includegraphics[width=1\linewidth, trim=1.5cm 0cm 0.7cm 0cm, clip]{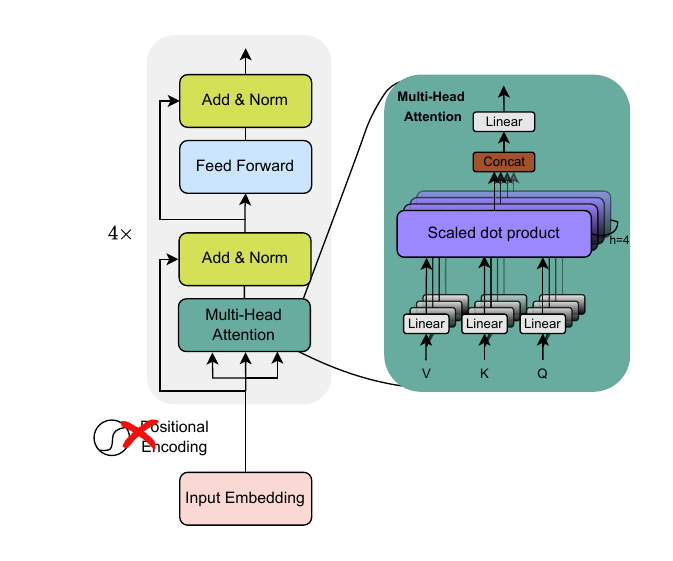}
    \caption{The transformer architecture used in this paper.}
    \label{fig:transformer-paper}
\end{figure}

As shown in figure \ref{fig:transformer-paper}, the core of our model comprises four transformer blocks, each equipped with four attention heads and a feedforward FC layer. The forward pass adheres to the standard transformer mechanism, incorporating layer normalization and residual connections. Positional embeddings are intentionally omitted as the state space does not require sequential information across the nodes. Two empty feature vectors, \( g_s \) and \( g_a \), are appended to the input set of vectors and will be used for state representation and allocation, respectively. This design yields an embedding tensor \( H = [h_1, h_2, \dots, h_N, h_a, h_s] \), where each \( h_i \) represents the corresponding embedding tensor with a dimensionality of \( d = 32 \) for each input node feature \( g_i \) in the input state \( s = [g_1, g_2, \dots, g_N] \). The embedding tensors \( h_a \) and \( h_s \) correspond to the two empty vectors \( g_a \) and \( g_s \).

We employ two FC modules that operate on the Transformer's embeddings: one for node scoring and the other for resource allocation. The node scoring module processes the node embeddings $H^\text{n} = [h_1, h_2, \dots, h_N]$ to produce a vector of scores $Z^{s} = [z^{s}_1, z^{s}_2, \dots, z^{s}_N]$, formalized as in $Z^{s} = f_{\text{scoring}}(H^\text{n})$, and subsequently passed through a softmax layer to generate a probability distribution. This distribution is used to sample the node $v$ with which we will establish a channel. Additionally, the allocation module processes over the allocation embedding $h_a$ and computes a vector of size $K$, where each element represents a score corresponding to a discrete capacity share $Z^{a} = [z^{a}_1, z^{a}_2, \dots, z^{a}_K]$ which can be expressed as $Z^{a} = f_{\text{allocation}}(h_a)$. A similar sampling procedure is applied to these scores, yielding the allocation share $c$ as the final output. 

Eventually, the agent's decision-making unit process is based on an actor-critic structure, in which, the actor comprises the node selection and resource allocation units and the critic employs an MLP structure over the state embedding vector $h_s$ to estimate state value. This structure is trained by the PPO algorithm to yield stable results \cite{ppo}.

\section{Network Analysis}\label{sec:analysis}

LN solution to Bitcoin's scalability challenges, along with its fee charging mechanisms, raises concerns regarding privacy, security, and the imperative of decentralization \cite{tikhomirov2020quantitative,pili2021topological}. Numerous studies have examined the trend toward centralization within the LN through the analysis of available snapshots \cite{camilo2023analyzing,martinazzi2019evolution,perik2022analysis}; some mention this trend poses several critical challenges, including security vulnerabilities, the potential for monopolistic behavior, liquidity bottlenecks, the risk of transaction censorship, and other related issues \cite{zabka2024centrality}, meanwhile others attributing this trend to revenue maximization strategies \cite{impact} and the importance of proximity to network hubs \cite{hubs}. These findings highlight the need to address centralization from an agent-based perspective. In the following section, we detail our efforts to simulate the evolution of the LN using revenue-driven agents, developed as solutions to the JCNSRA problem.

We utilized three of the models we developed from the JCNSRA problem. Each model acts based on different localizations: 50, 100, and 200 nodes. The training scheme of these models includes localized versions randomly chosen from various parts of the main LN graph. Hence, each agent can be utilized simultaneously in many different localized versions, as a separate agent. 

For each setup, we analyze the variations in four widely-used centrality measures: degree centrality \cite{degree-centrality}, betweenness centrality \cite{betweenness-centrality}, eigenvector centrality \cite{eigenvector-centrality}, and closeness centrality \cite{closeness-centrality}. The degree centrality for a node is the fraction of nodes it is connected to. Betweenness centrality assesses the importance of a node based on the number of shortest paths passing through it, while eigenvector centrality evaluates a node's influence by considering the significance of its connections. Closeness centrality quantifies how efficiently a node can reach all other nodes using the shortest paths. Each measure captures a different aspect of network centralization. Our focus primarily lies on betweenness and closeness centrality, as they closely correspond to the network's revenue potential and compactness, respectively \cite{how-to-profit, lightning-gym}.

In section \ref{sec:results}, we plotted histograms of the centrality distributions to evaluate the feasibility of achieving uniformity, which signifies a more equitable distribution of centrality and, consequently, a higher level of decentralization within LN. However, upon examination, it became evident that solely interpreting these histograms can be challenging. Consequently, we employed numerical metrics to evaluate these differences with greater accuracy.

We employed three distinct entropy and diversity metrics to compare the centralization of the base snapshot with our evolved versions. Shannon entropy and Rényi entropy \cite{bromiley2004shannon} were used to measure the randomness of distributions, with Rényi entropy offering increased sensitivity to hubs and centralized entities and the Gini index \cite{gastwirth1972estimation} quantifies inequality within a distribution. Ideally, the goal is to observe a decrease in the Gini index alongside an increase in entropy metrics.

Finally, we assess the modularity of both the LN and the evolved network. Network modularity is a measure of the strength of the division of a network into modules or communities, quantifying how well a network can be partitioned into groups where connections are denser within groups than between them.\cite{ye2012new} First, we identify the partition with the highest modularity for each network using the Louvain algorithm \cite{de2011generalized}. We then calculate the modularity corresponding to these partitions. 

\section{Results}\label{sec:results}

\subsection{Experimental Setup}
We perform experiments in three different localized settings. We sample three connected graphs with 50, 100, and 200 number of nodes. In each setting, the agent is set to open 5, 10, or 15 channels respectively. These numbers are set to approximate a reasonable number of channels for a profit-maximizing participant in the real LN.

As discussed in the section \ref{sec:approach}, we discretize the allocation action space into $K = 10$ parts. Furthermore, the total allocation budget is set to $10^7$ Satoshis. This value approximates a reasonable total budget of a node with regard to transaction amounts, counts, and the predefined number of channels to be opened. 

Similar to DyFEn \cite{Aida-Kiana}, we limit the possible transaction amounts to three distinct values of 10000, 50000, and 100000 Satoshis, with each value being generated 200 times in each step. Derived from the localized graph, we hence create three sub-graphs dynamically in each step with the channels capable of passing through the corresponding amount.

\subsection{Baselines}
We implement and evaluate several baseline heuristics and models. Our experiments include training and assessing a deep MLP, a GNN, and a hybrid model combining the GNN with our proposed transformer. Additionally, we benchmark five key node selection heuristics, which encompass random selection and top-k or bottom-k selection based on betweenness centrality or degree centrality. In all these heuristic methods, resources are uniformly allocated.

\textbf{MLP: }
We employed the default architecture of the stable-baselines3 (SB3) \cite{sb3} MLP network, modifying the number of layers in both the actor and critic networks. Specifically, we used a deep MLP with four hidden layers, each containing 128 neurons, and a lighter MLP with two hidden layers, each containing 256 neurons, for the critic network. Despite testing multiple configurations, the results remained consistent, indicating the ineffectiveness of this module in handling the complexity of the task.

\textbf{GNN: }
To ensure compatibility of GNN-based models with our system, we utilized specific classes from a modified open-source version of the SB3 repository\footnote{https://github.com/BlueBug12/custom\_stable\_baselines}. This adaptation was necessary to make our observation space compatible with the SB3 PPO algorithm, which does not natively support graph-based observations. In place of our transformer module, we implemented a two-layer Graph Convolutional Network (GCN) with a hidden size of $d = 64$ neurons per layer, while retaining the rest of our original architecture. We experimented with various GNN models, including GraphSAGE \cite{velivckovic2017graph}, GAT \cite{hamilton2017inductive}, and GATv2 \cite{gatv2}, as well as different pooling strategies such as max pooling, min pooling, and mean pooling. The optimal configuration was found to be the GCN model combined with mean pooling.

\textbf{GNN+Transformer: }
We explored a hybrid architecture by connecting the two-layer GCN described above to our proposed transformer module. In this approach, the input graph data was first processed by the GCN to generate embeddings for each node, which were then fed into the Transformer. We hypothesized that this combination would enhance performance by better capturing the underlying graph structure. However, the resulting architecture proved challenging to optimize, and we were unable to replicate or surpass the results achieved with the Transformer alone.

\textbf{Transformer-Node-Selector: }
To assess the performance of our proposed transformer model in resource allocation, we conducted a comparative evaluation by utilizing our original transformer model for node selection but substituting the model's resource allocation mechanism with a static, uniform distribution across each episode. While uniform allocation may theoretically present itself as a viable baseline strategy, we hypothesize that the ability to deviate from this naive approach and employ more informed, adaptive allocations will demonstrate performance advantages.

\subsection{Heuristics}
We employed several well-established heuristics commonly referenced in the literature for node attachment strategies. As these heuristics do not specify an approach for resource allocation, we applied a uniform allocation strategy across all of them to ensure a consistent basis for comparison with our model and other baseline methods.

\textbf{Random: } The random strategy is our simplest heuristic. This strategy would select our nodes for connection uniformly from the sampled graph. This strategy can be quickly computed, and although its primary use is to act as a comparative benchmark, it mitigates centralization by not favoring any specific connection point.

\textbf{Degree-Based: }
The Top-k-Degree strategy ranks nodes within the sampled graph by degree, selecting the top k nodes for our proposed node-selection solution. This approach efficiently constructs the candidate set, as each node’s degree can be quickly retrieved from adjacency lists, and sorting nodes by degree is computationally manageable.

Connecting to nodes with the highest degrees exemplifies a strong form of preferential attachment, known to foster a “rich-get-richer” phenomenon characteristic of scale-free networks \cite{impact, barabasi1999emergence}. This strategy is likely a contributing factor to the highly centralized structures observed in the LN today. Indeed, Top-k-Degree selection methods were previously implemented in LND’s autopilot feature and have since sparked significant discourse within the community \cite{pickhardt2018barabasi}.

However, considering the challenges posed by centralization and the potential benefits of connection to less-central nodes for increased revenue, we propose an alternative Bottom-k-Degree heuristic and test this hypothesis by comparing its performance to the Top-k-Degree method.

\textbf{Betweenness-Based: } 
Betweenness centrality \cite{betweenness-centrality} is a key concept in network analysis that quantifies the influence of a node within a graph based on its position relative to other nodes. Mathematically, the betweenness centrality  $ C_B(v)$ of a vertex $v$ is defined as:

\begin{equation}
    C_B(v) = \sum_{s \neq v \neq t} \frac{\sigma_{st}(v)}{\sigma_{st}}
\end{equation}

where:
\begin{itemize}
    \item $ \sigma_{st} $ is the total number of shortest paths from node $ s $ to node $ t $,
    \item $ \sigma_{st}(v) $ is the number of those paths that pass through vertex $ v $.
\end{itemize}

This measure reflects how often a node acts as a bridge along the shortest paths between other nodes in the network. A higher betweenness centrality indicates that a node has greater potential to control communication and flow within the network, making it crucial for understanding the structure and dynamics of complex systems.

In the LN context, nodes with high betweenness centrality frequently serve as intermediaries in the weight-based routing algorithm, appearing in multiple payment paths. Because these nodes enable broad network access with minimal routing fees, they are strong candidates for node attachment. Notably, connections through these nodes also tend to produce shorter payment paths, enhancing transaction reliability \cite{impact}.

Following this reasoning, the Top-k-Betweenness attachment strategy selects the top k nodes with the highest betweenness centrality values. Furthermore, inspired by the Degree-based Bottom-k-Degree approach, we also propose a Bottom-k-Betweenness heuristic alongside the Top-k-Betweenness strategy to explore alternative attachment dynamics.

\begin{figure*}[htbp!]
     \centering
        {\includegraphics[width=1\textwidth, keepaspectratio]{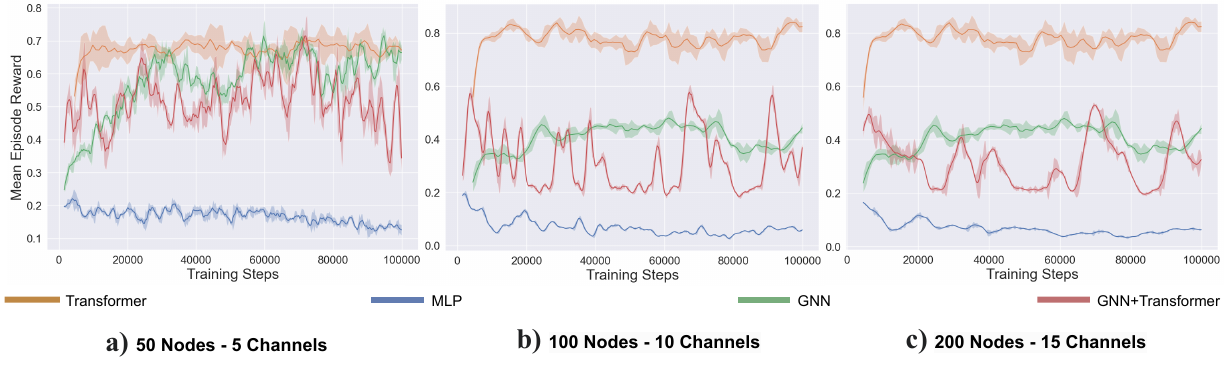}}
        \caption{Mean training reward plot for the proposed model and baselines in the three experimental settings.}

        \label{fig:plots}
\end{figure*}

\subsection{Evaluation Metric}
For each experimental setting, we independently deploy each model on a set of one thousand randomly sampled graphs. The reward for each episode is determined by the revenue generated in the final step and is normalized by a constant. The average reward attributed to each model in a given setting is then computed by averaging the rewards across all sampled graphs.

\subsection{Performance Evaluation}
Table \ref{tab:Evaluation} provides a detailed analysis of each model's performance across various settings. Contrary to prevailing maneuvers in the literature on the top-k method \cite{impact}, our results indicate a significant outperformance by bottom-k methods. This observation aligns with the understanding that connecting to high-centrality nodes does not necessarily ensure high income for the agent node. Additionally, we hypothesize that bottom-k measures enhance closeness between distant nodes within the network, yielding the shortest path for a greater number of transactions.

As depicted in Table \ref{tab:Evaluation}, our proposed model demonstrates clear superiority in overall performance among other models and heuristics. Figure \ref{fig:plots} also indicates a significantly faster convergence speed of this method in comparison to other proposed models. 

The MLP consistently demonstrates the lowest performance across all configurations, with a significant gap distinguishing it from other models. This marked disparity suggests a fundamental mismatch between the problem requirements and the MLP architecture. Likely contributing factors include the complexity of the task and the need to preserve node positional information during computation, which the fully connected structure of MLPs is inherently ill-suited to handle.

The GNN model exhibits competitive performance in the 50-node configuration. However, as subgraph size increases, this model lags behind the proposed Transformer model. This discrepancy may be attributed to the inherent capacity of transformers to manage more extensive dependencies. The GNN+Transformer architecture reflects behavioral traits from both its GNN and Transformer components, but its relatively lower performance may highlight challenges in harmonizing these elements. Specifically, the complex representation space generated by the GNN could potentially destabilize the input space required by the Transformer. These observations, along with the fact that GNNs are traditionally optimized for graph-based tasks, suggest a need for further exploration of GNN-based architectures tailored to this problem.

Heuristic approaches offer valuable insights into this problem. Notably, the Top-k methods perform the worst, suggesting that maximizing revenue incentives do not align with proximity to high-impact nodes. In contrast, the Bottom-k metrics outperform all other heuristics and baselines, excluding the Transformer-Node-Selector, supporting the notion that connections to less central nodes can be advantageous. Additionally, Top-k-Betweenness outperforms Top-k-Degree, while Bottom-k-Degree surpasses Bottom-k-Betweenness. This result suggests a potentially better strategy that combines both measures instead of using each one separately. This further justifies our inclusion of both degree centrality and transaction flow, which serves as a proxy for betweenness centrality, in our state space.

\begin{table}[ht]
\centering
\caption{Average normalized revenue gained through each setting.}
\setlength{\tabcolsep}{2pt} 
\begin{tabular}{@{}lccc@{}}
\toprule
\multirow{2}{*}{\textbf{Model}} & &\textbf{Nodes - Channels}& \\
\cdashline{2-4} 
\\
{} & \textbf{50-5}& \textbf{100-10} & \textbf{200-15}\\
\midrule
Transformer-Node-Selector & 0.6467 &  0.7789 & 0.7854\\
GNN+Transformer & 0.2596 &  0.5359 & 0.3786\\
GNN & 0.6483 & 0.5022 & 0.4245\\
MLP & 0.1448 & 0.0764 & 0.0499\\
\hdashline
Random & 0.2138 & 0.2597 & 0.2162\\
Bottom-k-Betweenness & 0.5053 & 0.6445 & 0.7338\\
Top-k-Betweenness & 0.1441 & 0.1700 & 0.1492\\
Bottom-k-Degree & 0.5446 & 0.6734 & 0.7472\\
Top-k-Degree & 0.1149  & 0.1257 & 0.1097\\
\midrule
Transformer & \textbf{0.6798} & \textbf{0.8081} & \textbf{0.8171}\\
\bottomrule
\end{tabular}
\label{tab:Evaluation}
\end{table}

\subsection{Resource Allocation Evaluation}

We evaluated the effectiveness of our resource allocation module by benchmarking it against the Transformer-Node-Selector baseline. As demonstrated in Table~\ref{tab:Evaluation}, our main transformer-based approach outperforms the baseline, underscoring the robustness of our resource allocation strategy. This finding highlights the module's capability to derive more effective allocation strategies compared to a uniform policy, which serves as a reasonable yet naive alternative \cite{van2019designing}. To further validate the superiority of our approach, we conducted statistical evaluations using a t-test. The results, detailed in Table~\ref{tab:ttest}, indicate that the observed differences in mean rewards between the two models are statistically significant, supported by low p-values derived from 1,000 samples. Due to the markedly lower performance of other baselines relative to the Transformer-Node-Selector model, additional statistical tests were deemed unnecessary.

\begin{figure*}[htbp!]
     \centering
        {\includegraphics[width=1\textwidth, keepaspectratio]{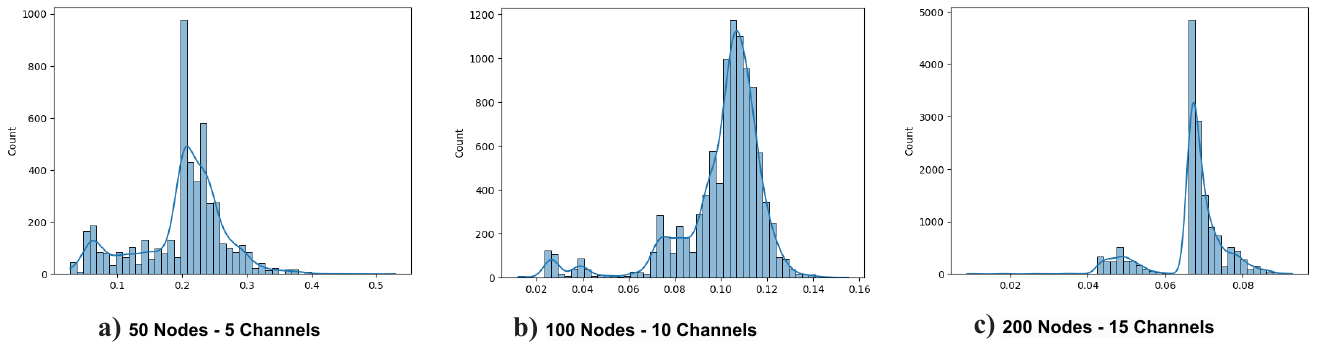}}
        \caption{Distribution of allocated resources by the proposed model in each experimental setting.}

        \label{fig:dists}
\end{figure*}

We also present the allocation distributions of our resource allocation module in Figure~\ref{fig:dists}. The results reveal a noticeable deviation from the uniform policy under the 100-10 setting, with smaller deviations observed in the 50-5 and 200-15 configurations. Despite these differences, all allocation strategies derived by our module consistently outperform the naive uniform distribution, as evidenced by the results in Table~\ref{tab:ttest}. These findings provide robust support for our hypothesis that smart deviations from uniform allocation can lead to more effective resource utilization and improved outcomes.

It is also noteworthy to consider the efficiency of the uniform allocation strategy. The observed similarity to a uniform distribution in the 200-15 setting reflects the increased complexity of the JCNSRA problem as both sample size and long-term dependencies grow. Conversely, the 50-5 setting indicates reduced similarity among network participants, suggesting that larger sampled networks tend to exhibit greater heterogeneity.

\begin{table}[ht]
\centering
\caption{Statistical T-Test between Transformer and Transformer-Node-Selector.}
\setlength{\tabcolsep}{4pt} 
\renewcommand{\arraystretch}{1} 
\begin{tabular}{@{}lcccc@{}}
\toprule

\textbf{Setting}&\textbf{Model} & \textbf{Mean}& \textbf{STD} & \textbf{P-Value}\\
\midrule

\multirow{3}{*} {50-5}& Transformer & 0.6798 & 0.3845 & \multirow{3}{*}{0.0231}\\
\cdashline{2-4} \\
{}& Transformer-Node-Selector & 0.6467 & 0.3579 & {}\\

\midrule
\multirow{3}{*} {100-10}& Transformer & 0.8081 & 0.3503 &  \multirow{3}{*}{0.0324}\\
\cdashline{2-4} 
\\
{}& Transformer-Node-Selector & 0.7789 & 0.3561 & {}\\
\midrule
\multirow{3}{*} {200-15}& Transformer & 0.8171 & 0.3657 &  \multirow{3}{*}{0.0218}\\
\cdashline{2-4} 
\\
{}& Transformer-Node-Selector & 0.7854 & 0.3353 & {}\\

\bottomrule
\end{tabular}
\label{tab:ttest}
\end{table}

\subsection{Network Analysis}

\begin{figure*}[htbp]
    \centering
    \includegraphics[width=\linewidth]{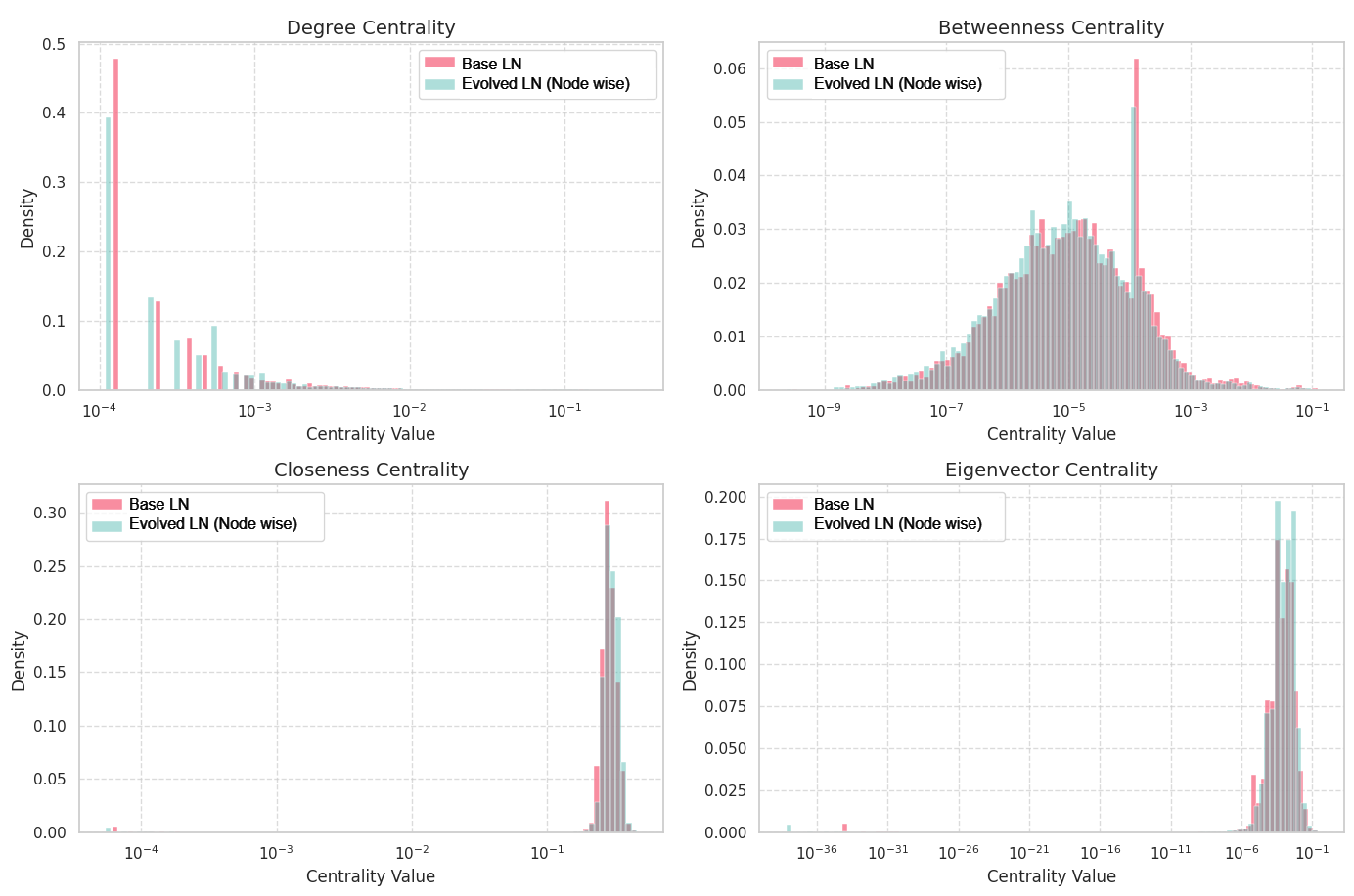}
    \caption{Distributions of centrality measures for the LN snapshot and evolved LN.}
    \label{fig:cent_plot}
\end{figure*}

We deployed our models over 2,000 episodes, randomly selecting each agent per episode with probabilities designed to maintain the ratio of additional channels and nodes consistent with the LN evolution trend. We calculate four different centralities for the LN snapshot and the evolved version. Figure \ref{fig:cent_plot} contains histograms of each measurement.

Our observations indicate that the evolved version demonstrated improvements across most centrality measures. Both Shannon and Rényi entropy metrics increased for nearly all measures, including Betweenness and Closeness, suggesting an increase in the similarity of nodes' centralities. Additionally, the rise in Eigenvector centrality entropy indicates that more nodes have become influential due to deploying our models. Although the decrease in Rényi entropy for Degree centrality suggests an increase in hub connectivity, this effect appears to be offset by more significant changes in the graph's lower degree and centrality regions. Lastly, the Gini index displayed inconsistent reductions. This could be attributed to the models' structural consistency. Each centrality-metric pair score is detailed in Table \ref{tab:Evolved}.

Our results due to modularities show that the modularity of the optimal partition for the LN decreases from $0.3970$ to $0.3754$ after deploying our agent. This reduction suggests that our approach effectively connects more modules within the graph.

\begin{table}[ht]
\centering
\caption{Centrality metric differences from Base LN graph analysis for different evolutions.}
\setlength{\tabcolsep}{7pt} 
\begin{tabular}{@{}lcccc@{}}
\toprule
\textbf{Centrality} & \textbf{Shannon} & \textbf{Rényi} & \textbf{Gini Index} \\
\midrule
Degree  & 0.1039 & -0.0406 & -0.0169 \\
Betweenness  & 0.1651 & 0.0773 & -0.0052 \\
Eigenvector  & 0.2877 & 0.2916 & -0.0459 \\
Closeness  & 0.1164 & 0.1165 & 0.0001 \\
\bottomrule
\end{tabular}
\label{tab:Evolved}
\end{table}

\section{Discussion}\label{sed:discussion}
This work offers a promising solution to the JCNSRA problem, focusing on profit maximization. Our method is readily compatible with new snapshots of the LN, provided they adhere to a specific format. This compatibility suggests that deployment on the LN would require minor modifications. However, given the financial sensitivity of this task, additional improvements may be necessary to ensure precision. For instance, the simulation process could be further refined by incorporating alternative routing mechanisms that have been discussed in the literature \cite{emergent}.

A key limitation of our work is the lack of advanced computational resources, which may have constrained the complexity of our models and the duration of the training phase. We acknowledge that utilizing more advanced architectures and extending the training period could lead to further improvements.

Moreover, Sampling plays a critical role in solving the JCNSRA problem within the LN. The LN exhibits a small-world architecture, with high clustering coefficients. This structure indicates that nodes are often part of tightly connected local sub-networks \cite{seres2020topological}. This might facilitate transactions within local regions for practical reasons, such as lower fees and reduced latency. Nevertheless, with revenue maximization as the primary objective and potential access to comprehensive network data, adopting a global perspective should yield the most effective resource allocation policies. Future work might aim to address the scalability challenges by exploring approaches that enable the agent to operate effectively on the entire LN graph.

It is also important to consider fee-setting strategies, as our approach currently utilizes constant fees based on the median across LN nodes. More sophisticated strategies could replace this constant fee approach to optimize fees in conjunction with our approach, thereby improving the potential for revenue maximization.

\section{Conclusion}
This paper presents a DRL framework, empowered by transformers, to tackle the intricate JCNSRA problem relevant to profit-making in the LN. By enhancing the flow dynamics in the simulation environment, we bridge the gap between currently employed models and the real-world dynamics of LN. The proposed model outperforms existing baselines. Finally, our network analysis examines the incentive structure of the LN and demonstrates that deploying our rational agents within the LN leads to a more decentralized topology. Contrary to the mixed opinions in the literature \cite{bertucci2020incentives, carotti2024rational}, our findings suggest that the network naturally tends towards decentralization, making it both decentralized by design and incentive-compatible.

\bibliographystyle{IEEEtran}
\bibliography{IEEEexample}

\end{document}